\documentclass{bmvc2k}

\usepackage{graphicx}
\usepackage{bbm}
\usepackage{tikz}
\usepackage{comment}
\usepackage{color}
\usepackage{caption}
\usepackage{subcaption}
\usepackage{booktabs}
\usepackage{amsmath,amssymb}
\usepackage{enumitem}
\usepackage{parskip}
\definecolor{red}{rgb}{1, 0, 0}
\usepackage{amsmath}
\makeatletter
\g@addto@macro\normalsize{%
  \setlength\abovedisplayskip{10pt}
  \setlength\belowdisplayskip{5pt}
  \setlength\abovedisplayshortskip{0pt}
  \setlength\belowdisplayshortskip{0pt}
}
\makeatother

\title{KPE: Keypoint Pose Encoding for Transformer-based Image Generation}

\addauthor{Soon Yau Cheong}{s.cheong@surrey.ac.uk}{1}
\addauthor{Armin Mustafa}{armin.mustafa@surrey.ac.uk}{1}
\addauthor{Andrew Gilbert}{a.gilbert@surrey.ac.uk}{1}

\addinstitution{
 University of Surrey, UK
}

\runninghead{Cheong, Mustafa, Gilbert}{Keypoint Pose Encoding for Image Generation}

\begin{document}

\maketitle
\vspace{-0.6cm}
\begin{abstract}
Transformers have recently been shown to generate high quality images from text input. However, the existing method of pose conditioning using skeleton image tokens is computationally inefficient and generate low quality images. Therefore we propose a new method; Keypoint Pose Encoding (KPE); KPE is 10$\times$ more memory efficient and over 73\% faster at generating high quality images from text input conditioned on the pose. The pose constraint improves the image quality and reduces errors on body extremities such as arms and legs. The additional benefits include invariance to changes in the target image domain and image resolution, making it easily scalable to higher resolution images. We demonstrate the versatility of KPE by generating photorealistic multiperson images derived from the DeepFashion dataset \cite{deepfashion}.We also introduce a evaluation method People Count Error (PCE) that is effective in detecting error in generated human images.

\end{abstract}
\begin{figure}[!htb]
\centering
    \begin{subfigure}{0.4\textwidth}
    \includegraphics[height=6cm]{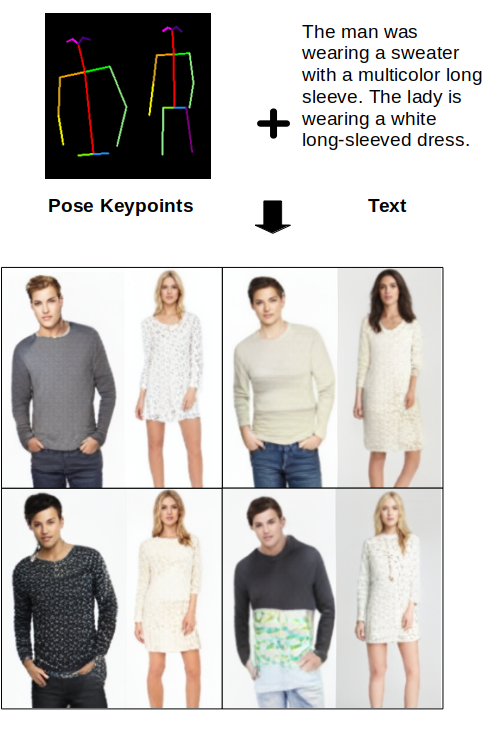}
    \caption{}
    \end{subfigure}
    \begin{subfigure}{0.5\textwidth}
    \includegraphics[height=6cm]{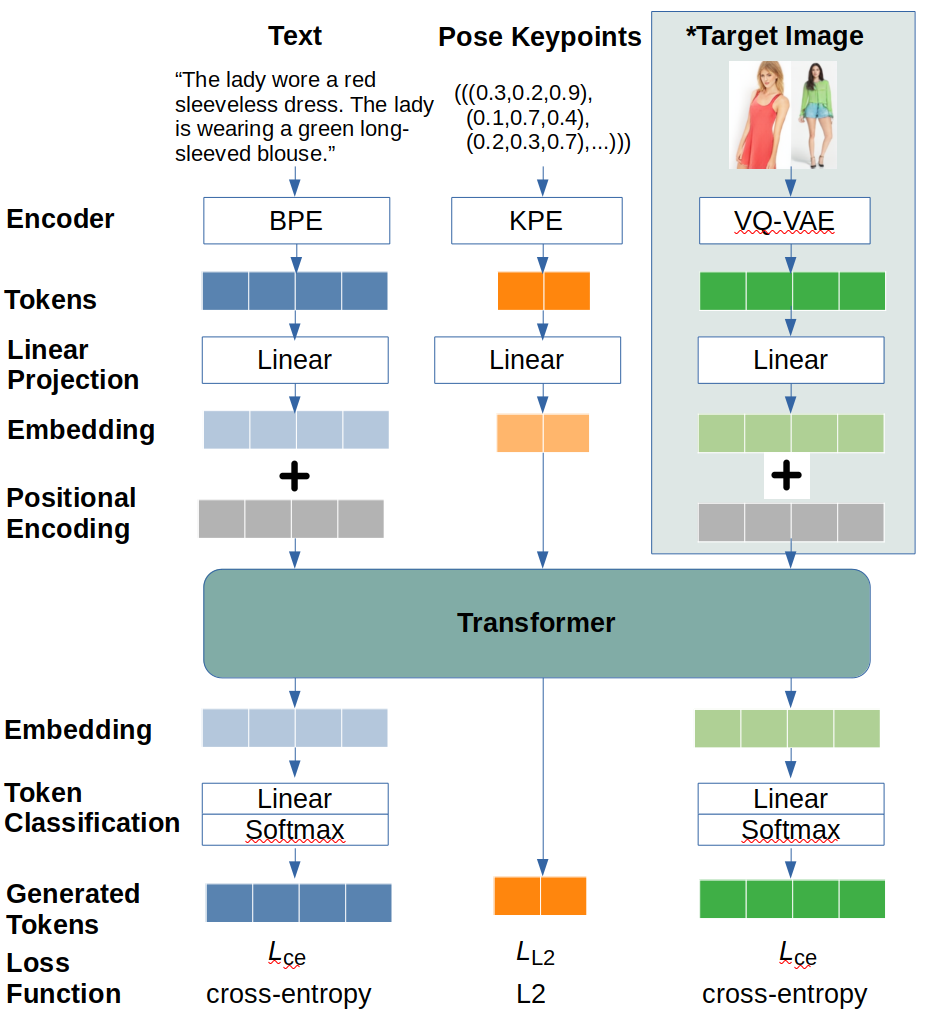}
    \caption{}
    \end{subfigure}

    \caption{(a) Our pose constrained text-to-image model supports partial and full pose view, multiple people, different genders, at different scales. (b) The Architectural diagram of our pose-guided text-to-image generation model. The text, pose keypoints and image are encoded into tokens and go into an transformer. *The target image encoding section is required only for training and is not needed in inference.}
\label{fig:Overview}
\end{figure}

\section{Introduction}
\vspace{-2mm}
Despite recent advancements in text-to-image generation, generating full-body multiperson images is still challenging due to the human body's high degree of freedom. It is difficult to describe in words entirely the body shape, size, pose, clothing details, position in the images and camera view. Hence in this paper we propose to add human pose as an additional input to text, to improve the accuracy and fidelity of people being generated through. We can see this as enforcing disentanglement of content and style of image \cite{styletransfer} where the content is the pose, and the text depicts the style. Concurrent to our work, Text2Human \cite{text2human} uses text and pose as input, but their method involves complex dedicated neural networks. At the same time, our approach is simple and generic to the transformer architecture. There are several families of text-to-image generation algorithms, including GANs and diffusion models and the focus of this paper is autoregressive transformer \cite{transformer} such as DALL-E \cite{dalle} which we use as a baseline in our study.

The standard method of representing pose for transformers, e.g. \cite{Makeascene} is to convert a skeleton image into discrete image tokens with a dVAE. This is the method adopted by VQGAN \cite{vqgan} which frames pose-to-image as an image-to-image problem. However, this image pose encoding has shortcomings, primarily with the high computational complexity, which increases quadratically with image resolution, limiting the output to low resolution. Therefore, we have devised \textbf{Keypoint Pose Encoding} (KPE) - a novel, efficient and accurate pose representation suitable for a transformer. Instead of using the high dimensional skeleton image mainly containing redundant information, we focus on only the body joint keypoints for pose representation. The low dimensional representation is invariant to changing the target image resolution or domain, e.g. from the natural landscape to synthetic objects. We show it to be 10x more memory efficient and increase computational inference speed by over $73\%$ in the experiment section. 

To measure success and motivated by the inadequacy of existing metrics to measure image errors specific to people, we devised the \textbf{People Count Error} (PCE) metric to measure the false positive rate when generating images of multiple people. Therefore, we can both empirically and qualitatively show that the disentanglement improves the fidelity of people generated with a significantly reduced number of false positive or erroneous body parts. 

In summary, our key contributions are:
\begin{enumerate}
    \item \textbf{Novel Keypoint Pose Encoding (KPE)} To enable the \emph{tokenisation} of a human pose representation that is computationally efficient and invariant to changes in target images such as the resolution. 
    \item \textbf{Introduction of a person centric metric} A new metric to measure the false error rate of generated humans in multiperson images in rendered images. This metric helps to illustrate that the disentanglement of pose and text leads to better higher quality results with reduced false positive humans.
    \item \textbf{Challenging person generation} Our pose constrained text-to-image model supports partial and complete pose views, multiple people, all at different scales.
\end{enumerate}

\section{Related Work}
\subsection{Text-to-Image Generation}
Most GAN-based text-to-image models such as StackGAN\cite{stackgan}, AttenGAN\cite{attngan}, DM-GAN\cite{dmgan}, DF-GAN\cite{dfgan} and XMC-GAN\cite{xmcgan} are forms of conditional GAN \cite{cgan} where the text sequence is projected to an embedding vector as a conditioning feature, i.e. the text is modelled as continuous variables. More recently, researchers turned their attention to the transformer architecture that has proven to excel in natural language processing (NLP) tasks. Instead of using a single embedding vector for the entire word sequence, it is encoded into a sequence of discrete text tokens. Each token is then projected into the transformer's embedding space. The transformer is trained to predict the text tokens autoregressively in sequence. 

However, there are challenges in applying the transformer to computer vision tasks. The high dimensionality of image pixels makes it computationally expensive as the transformer computational complexity is $O(N^2)$. Moreover, image pixels are considered continuous real values and do not fit into the discrete token needed by the transformer. DALL-E \cite{dalle} addressed these problems by using VQ-VAE\cite{vqvae} which is a variant of dVAE to encode the image into discrete tokens. Hence, both text and image are represented as discrete tokens, allowing the adoption of generic transformers to predict the image tokens from text tokens. The resulting image is generated by feeding the image tokens to the decoder part of dVAE. This method forms the basis of many modern transformer-based text-to-image models such as CogView\cite{cogview}, N\"{U}WA\cite{nuwa} and Make-a-scene\cite{Makeascene}. We employ a similar method to tokenise the text in our models but condition our transformer on a further additional modality, i.e., pose, to enrich the image quality and precision. 

Recently, a new image generation family known as diffusion models  \cite{Ho2020}\cite{Ho2021}\cite{Dhariwal2021} emerge. In training, noise is added to target image in forward diffusion steps, and a U-Net\cite{unet} is trained to predict the noise in reverse diffusion steps to generate images from noise. GLIDE\cite{glide}, DALL-E 2\cite{dalle2} and Imagen\cite{imagen} applies additional text condition as embedding into U-Net to create text-to-image models. Using our KPE method to create pose embedding has shown to work with diffusion models but the computational benefit is more profound for transformer hence our focus.

\subsection{Pose Guided Image Generation}
There are a few existing methods that represent pose in the context of image generation; including, body keypoint heatmaps \cite{Ma2017}\cite{Siarohin2018}\cite{Yang2020}, segmentation maps \cite{gaugan}\cite{Makeascene} and a skeleton image \cite{pix2pix}\cite{Zhu2019}. 
The pose-to-image is framed as image-to-image, where an input image is a form of 2D spatial tensor representing pose, and the output image is the person image. However, these representations include the redundant background in addition to the foreground segmentation or skeleton data.
The reason for using the 2D spatial tensor for pose representation is that the spatial information is required for convolutional layers in GANs \cite{Tompson2014}, but this is no longer a pre-requisite for transformer. Despite this, recent generative transformer models \cite{vqgan}\cite{nuwa}\cite{Makeascene} continue using an image for pose representation by encoding image into discrete image tokens. The image tokenisation process can be very slow, \cite{dalle_mini} attempts to reduce the training time by pre-encoding the images into tokens, but this prohibits the use of augmentation onto the images and poses during training. Thus, several papers \cite{skeletor} \cite{Li2021} realised the shortcoming of using skeleton images and started using keypoint for pose estimation regression. However, we are the first to use pose keypoint to guide the image generation. 

\section{Method}

Figure \ref{fig:Overview}(b) shows the overall architecture of our pose constrained text-to-image model. The first stage is to convert the text, pose keypoints and image into tokens with their respective encoders. Then the tokens are projected into an embedding space before adding positional encoding. We use learnable positional encoding for text tokens and axial positional encoding \cite{axial} for image tokens due to its 2D structure. We do not use positional encoding for keypoint tokens as they are equally important for all positions of the image tokens. The transformer is made of 12 transformer encoder blocks adopted from \cite{dalle_pytorch}, and there is no transformer decoder block. We will now describe the details of the token encoders.

\subsection{Text Encoder}
Like DALL-E \cite{dalle}, we use the BPE (Byte Pair Encoding) tokeniser \cite{bpe} for text tokenisation, the encoder breaks the word into subwords, and they are assigned discrete identifier numbers, which become the text tokens. The text tokens are then projected into embedding  $\mathbb{R}^{t \times d}$, where $t$ is the fixed text token input length and $d$ is the transformer dimension. The BPE tokeniser is pretrained with the vocabulary of the dataset. 

\subsection{Image Encoder}
In our model, we use VQ-VAE \cite{vqvae} from VQGAN \cite{vqgan} for image encoding and decoding. Its GAN training pipeline has produced better image quality than the dVAE used by \cite{dalle}. The encoder $E$ first converts the continuous image  $x \in \mathbb{R}^{H\times W \times C}$ with height $H$, width $W$ and colour channel $C$ into code  $\hat{z} = E(x) \in \mathbb{R}^{h\times w \times n_z}$ with $h$ and $w$ the spatial dimension and $n_z$ the dimension of the code. Then, each of the codes is quantised \textbf{q}(.) to its closest discrete codebook entry $z_k$ using the equation:
\vspace{-4mm}
\begin{align}
    z_q = \textbf{q}(\hat{z}) := \left(\arg \min_{z_k \in Z} \|\hat{z}_{ij} - z_k \|\right) \in \mathbb{R}^{h\times w \times n_z}
    \label{eq:vqgan_quantise}
\end{align}

where $Z = \{z_k\}^K_{k=1} \in \mathbb{R}^{n_z}$ is the discrete codebook. The decoder G reconstructs the discrete code into image $\hat{x}$:
\vspace{-2mm}
\begin{align}
    \hat{x} = G(z_q) = G(\textbf{q}(E(x)))
\end{align}
The model and codebook can be trained end-to-end using the loss function:
\begin{align}
    L_{VQ}(E, G, Z) = \|x-\hat{x} \|^2 & + \|sg[E(x)-z_q] \|^2_2  \\ 
                                       & + \|sg[z_q]-E(x) \|^2_2 
\end{align}
where $sg[.]$ denotes stop-gradient operation.

Overall, the image is tokenised into $h \times w $ grid of discrete image tokens. Although $h$ and $w$ are hyperparameters, they have a linear correlation with image resolution $H$ and $W$ to maintain the same quality of image texture details. Therefore, increasing image resolution will lead to more extended image token lengths, resulting in a quadratic increase in computational complexity. The image encoder is pretrained with the target images. The discrete image tokens are then projected into transformer embedding $\mathbb{R}^{h \times w \times d}$.

\subsection{Keypoint Pose Encoder (KPE)}
\vspace{-0.2cm}
KPE is our method for pose representation. It converts pose positions for multiple people into keypoint tokens and then encodes them into a keypoint embedding.

A single 2D keypoint is defined as a tuple of $(x, y, v)$ where $x$ and $y$ are the normalised Cartesian coordinates in [0, 1] and $v$ is the visibility score in $[0, 1]$. We denote multiperson 2D keypoint format as $(x, y, v)_{i,j}$ where $i$ is the person index, $j$ is the keypoint index from 0 to $N-1$ where $N$ is the total number of keypoint defined. Different pose estimation models use different keypoint schemes, but the common keypoints are the nose, neck, shoulder, elbow, wrist, hip, knee, ankle, eye, ear, big toe, and heel. 

\begin{figure}[!htb]
\centering
\includegraphics[width=0.7\linewidth]{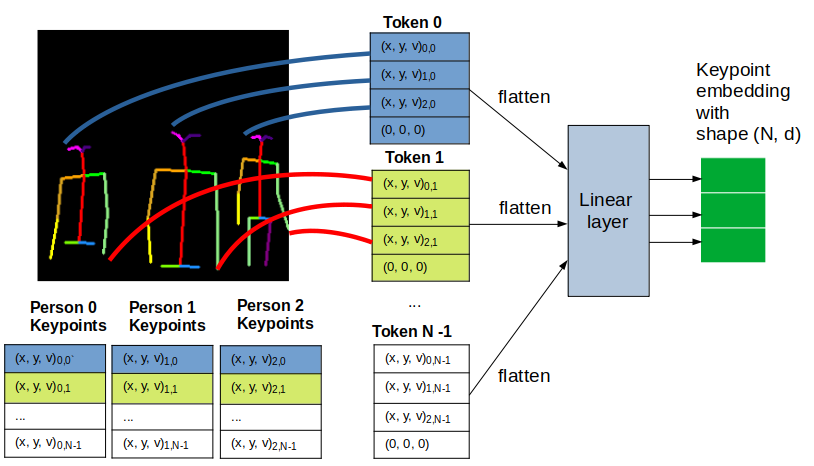}

\caption{The Block diagram showing KPE encoding multiperson pose to keypoint tokens. The tokens are flattened and projected into keypoint embedding within the transformer. The skeleton image is for illustration purposes as we use the keypoints directly from the pose estimation model's outputs.}
\label{fig:kpe}
\end{figure}

Figure \ref{fig:kpe} shows the process of KPE. The process is similar to the method in Vision Transformer (ViT)\cite{vit} of grouping image pixels into image patches, except that we group keypoints into keypoint tokens. Each keypoint token corresponds to a skeleton joint; in this illustration, keypoint token $0$ is designated for the right eye and keypoint token $1$ is for the left wrist. The right eye keypoint for all the people in the image are stacked together and moved into token $0$. In this example, we define the system to support up to four people, and if there are fewer than the maximum number of people, they will be padded with zeros in the keypoints. The same goes for keypoints that are not visible. The resulting keypoint tokens will have a length of $N$. In other words, the number of fixed with the keypoint definition does not change with the number of people in the image or changes in image resolution. It is worth noting that the keypoint token is not discrete; it is continuous in the range [0, 1].

The next step is to ensure that the pose embedding has the same dimension as the transformer embedding. We propose two methods; the first one is to pad the keypoint tokens with zeroes to match the transformer dimension. This method is the fastest as it does not require any arithmetic computation. It can accommodate many people up to constraint within $3 M <= d$ where $M$ is the maximum number of people, and $d$ is the transformer dimension.  We tested this method to be working faster. However, we use a linear layer to project the keypoint tokens into keypoint embedding for the generality of the unbounded number of people. Overall, the KPE converts multi-person keypoints $ \mathbb{R}^{M \times N \times 3}$  into embedding $ \mathbb{R}^{N \times d}$. 

\subsection{Training}
To train, the text tokens $T$, keypoint tokens $K$ and image tokens $I$ are concatenated to be fed into the transformer. The transformer output has the same length as the input, aiming to generate the same tokens as the input tokens. As the text tokens and image tokens are discrete, prediction of them becomes a multiclass classification problem, and we use the cross entropy loss $\mathcal{L}_{ce}$ as is common with transformer training. However, the keypoint tokens are continuous values, and we use an $L2$ loss $\mathcal{L}_{L2}$ on the keypoint embedding. Therefore, the overall loss function is: 
\vspace{-5mm}
\begin{equation}
    \mathcal{L} = \mathcal{L}_{ce}(T) + \lambda_{I}\mathcal{L}_{ce}(I) +  \lambda_{K}\mathcal{L}_{L2}(K) 
    \label{eq:kpe}
\end{equation}
$\lambda_{I}$ and $\lambda_{K} $ are constants. Higher value encourages more accurate people image and poses, respectively, and their values are discussed in Section~\ref{sec:implemantion}. 

\subsection{Inference}
We do not use input image inference. First, we fix the text and pose tokens and run a forward pass to generate score for image token. If we select the image token with the highest probability, the model will generate the same image every time and become deterministic. Therefore, to introduce variation in generating images, we select several tokens with the highest probability and sample them uniformly. The size of the sampling pool is a hyperparameter; setting it too low can give more variation but may also introduce more errors in the image. The sampled image token is concatenated with text and pose tokens to generate the next image token. The process repeats until all image tokens have been generated, then they will be decoded to generate an image.

\section{Experiments}
In this paper, we implement our KPE model and train it on a multiperson dataset derived from the DeepFashion \cite{deepfashion} dataset. In addition, we implemented two baseline models for comparison. DALL-E \cite{dalle} as an ablation study to understand the effect of using pose guidance on generated image quality. Moreover, we implement a DALL-E\cite{dalle}+VQGAN \cite{vqgan} model to compare the performance of KPE against VQGAN's method of using skeleton images for pose representation. We highlight the advantages of KPE in Section \ref{sec:dalle_vqgan}.

\subsection{DeepFashion Dataset}
We use DeepFashion's fashion synthesis benchmark dataset for the experiments. The original dataset contains 78.5K images of a single person and a brief description of the gender, clothing colour, and type. We derived a multiperson dataset by randomly sampling the single-person images, randomly resized them by 10\%, cropped them, and concatenated them into a single 256 $\times$ 256 image. The background of the individual images is not removed before the concatenation. The new images have between 1 to 3 non-overlapping people in various locations, sizes and poses. Then we use the OpenPose \cite{openpose} pose estimation approach to obtain the multiperson pose keypoints.  We will release code to enable the generation of the multiperson training and test data and splits.

\subsection{Evaluation Metrics}
We evaluate the performance of our approach on two aspects: the similarity or faithfulness of the generated images to text description and pose; and the image quality, which includes the realism of people.

\subsubsection{Similarity}
We use Object Keypoint Similarity (\textbf{OKS}) from the MSCOCO keypoint challenge \cite{oks} for keypoint accuracy. We also use Structural Similarity (SSIM) to compare the similarity of generated images to the reference images. Like \cite{Ma2017} we mask out the background, but we also we crop away the excessive background to avoid SSIM being dominated by the background to give the metric \textbf{Mask-SSIM}. We also use \textbf{CLIPSIM} \cite{godiva} to measure the similarity between text inputs and generated images.

\subsubsection{People Unrealism}
 To evaluate the \emph{realism} of the images, we use the perceptually trained Inception Score (\textbf{IS}) \cite{is} and \textbf{FID} \cite{fid}, in line with literature image generative models \cite{Ma2017}\cite{Zhu2019}\cite{dalle}\cite{nuwa}\cite{cogview}. However, as suggested by \cite{is_note}, IS sub-optimal, and we found that it is not good at measuring people-centric errors such as missing or extra limbs. To overcome this limitation, we propose a new evaluation metric People Count Error (\textbf{PCE}) to measure the unrealism of the people.  Given an image of people $x$, $gt()$ indicates the ground truth function that returns the labelled number of people in the image, and $h()$ is the function that returns the number of people detected by the pose estimation algorithm e.g. OpenPose \cite{openpose}. Therefore, PCE is defined as:
\begin{align}
    PCE(x) = 
    \begin{cases}
    1, & \text{if } h(x)\ne gt(x)\\
    0, & \text{otherwise}
    \end{cases}
\end{align}

PCE makes use of the rich body anatomy knowledge embodied in OpenPose. If a person in a erroneous generated image has three arms, OpenPose know human only has two arms, and therefore it will assume the third arm belong to another person, hence adding the person count. This discrepancy in people count is flagged by PCE as 1 (contain error). Visual examples of PCE are shown in the results in Fig \ref{fig:error} and further discussed in Section \ref{sec:error}. Moreover, unlike IS/FID  which requires a large amount of data, PCE applies to a single image, making it usable to find an error in an individual image. 

\subsection{Implementation details}
\label{sec:implemantion}
We adopt a two-stage training process like \cite{dalle}. The first stage trains VQ-VAE using the VQGAN pipeline on the target images to encode $256 \times 256$ images into $16 \times 16 = 256$ image tokens, where each token can assume 8192 possibilities. To compare against \cite{dalle}, we use an open-source implementation \cite{dalle_pytorch} with a transformer dimension $d$ of 512, with 8 heads and a depth of 12 encoder blocks. The text token length is 256, and the input text tokens will be truncated if they exceed this length. We train using the loss function in Equation \ref{eq:kpe} using OpenPose's BODY\_25 \cite{body25} pose format. Therefore, the keypoint token length is 25, corresponding to the 25 keypoints. 

The DALL-E+VQGAN model is also a text-and-pose guided model. The difference with our KPE model is that it uses VQGAN's pose representation method of using VQ-VAE to encode skeleton images into pose image tokens. Like VQGAN, we reuse the VQ-VAE pretrained on target images. The DALL-E+VQGAN's loss function is:
\begin{align}
    \mathcal{L} = \mathcal{L}_{ce}(T) + \lambda_{I}\mathcal{L}_{ce}(I) +  \lambda_{K}\mathcal{L}_{ce}(P)
\end{align}
where $P$ are the pose image tokens. 

We use the same VQ-VAE, training configuration and hyperparameters for all three models. Therefore, we chose a smaller VQ-VAE, which may not produce the most visually pleasing image quality, but this presents a fair comparison and ablation study. For loss constants, We use $\lambda_{I}$=7 from \cite{dalle_pytorch}. We tried 1 and 10 for $\lambda_{K}$ but did not notice much difference in the results. Eventually, we select $\lambda_{K}$=10 to have the same order of magnitude as $\lambda_{I}$. For the optimiser, we use Adam \cite{adam}, with initial learning rate of 0.0001, $\beta_{1}$=0.9, $\beta_{2}$=0.999. The learning rate is reduced by half if the loss has plateaued for 12 epochs until it reaches 1e-6. We use a batch size of 10 and train for 100 epochs on an RTX5000 GPU with 16GB GPU memory.

Due to the non-deterministic nature of image generation, we compute 5 images per sample in the test dataset and obtain the mean value of the metrics for the qualitative result. We sample image tokens from the top 0.1\% highest probability or 8 out of 8192 tokens. This sampling improves the image quality and consistency of metrics values.

\section{Results}
\begin{figure}[!htb]
    \begin{subfigure}{0.95\textwidth}
    \includegraphics[width=1\textwidth]{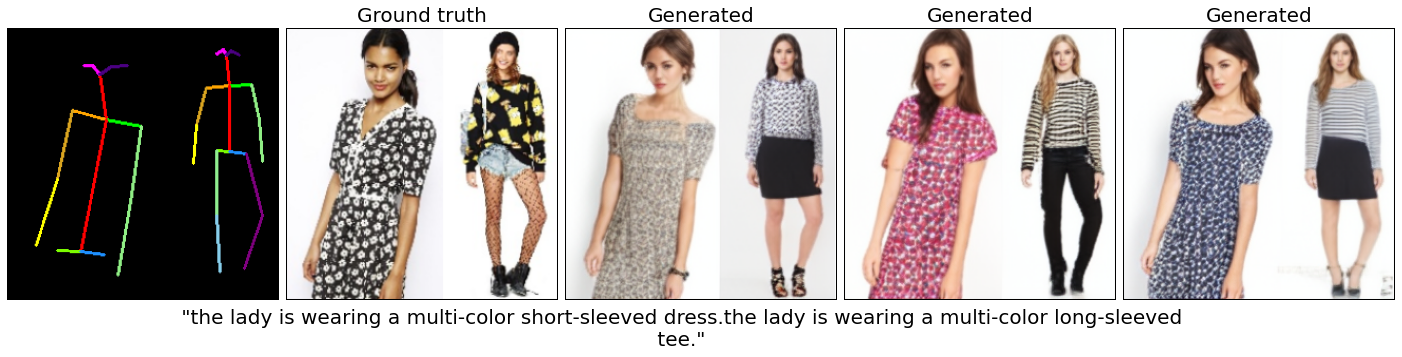}
    \vspace{-7mm}
    \caption{KPE model could generate multiperson with multiple scales. This example shows it works with a partial pose where the knee keypoints are missing.}
    \end{subfigure}

    \begin{subfigure}{0.95\textwidth}
    \centering
    \includegraphics[width=1\textwidth]{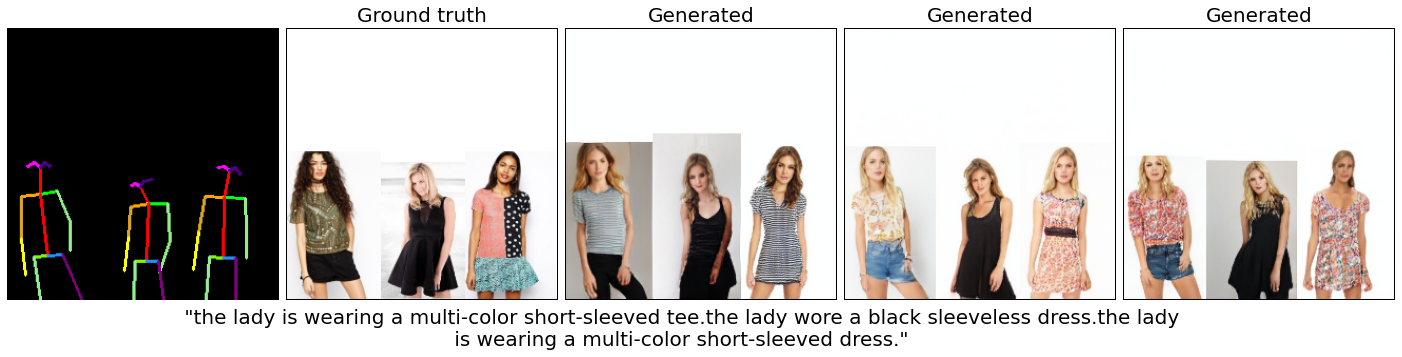}
    \vspace{-7mm}
    \caption{A variety of different poses by three people. Each of the people matches the text description in gender and clothing appearance.}
    \end{subfigure}
\vspace{3mm}    
\caption{KPE model can generate photorealistic people with an accurate pose. This figure shows the pose illustration, ground truth, and three generated samples. The ground truth images are not used in the inference, they are included only for comparison.}
\vspace{1cm}
\label{fig:deepfashion_1}
\end{figure}

\begin{center}
\begin{table}
\centering
\begin{tabular}{|c | c | c| c| c} 
\toprule
  Pose Method &  \cite{dalle} &  \cite{dalle}+ \cite{vqgan} & KPE (Proposed) \\ [0.5ex] 
\midrule
 Number of pose tokens 	$\downarrow$& - & 256 & \textbf{25} \\
 Relative inference speed $\uparrow$& \textbf{1.73$\times$} & 1.0$\times$ & \textbf{1.73$\times$} \\
 \midrule
 FID 	$\downarrow$ & 22.11 & 21.81 &  \textbf{20.39} \\
 PCE ($\times 10^{-3}$) $\downarrow$ & 8.2 & 1.2 & \textbf{0.6} \\ 
 CLIPSIM 	$\uparrow$ & \textbf{0.27} & \textbf{0.27} &  \textbf{0.27} \\
 IS 	$\uparrow$ & 2.912 & 3.027 &  \textbf{3.034} \\
 OKS  $\uparrow$& 0.598 & \textbf{0.970}  & \textbf{0.970} \\
 Mask-SSIM  $\uparrow$& 0.265 & 0.420  & \textbf{0.424} \\[0.5ex] 
\bottomrule
\end{tabular}
\vspace{2mm} 
\caption{Evaluation of different models on DeepFashion multiperson dataset. Our method, KPE, achieves the highest scores in all metrics.}
\label{table:deepfashion}
\end{table}
\end{center}
Table \ref{table:deepfashion} shows that our proposed method, KPE tops all the evaluation metrics; it achieves the highest FID and IS scores, indicating that KPE can generate realistic looking people. Figure \ref{fig:deepfashion_1} and \ref{fig:Overview}(a) show examples with various genders and quantity of people, with different scales, poses and occluded poses with missing keypoints. Given the OKS score of 0.97, which indicates a highly accurate pose, the high Mask-SSIM score suggests the generated images have gender and clothing appearance matching the text description. The CLIPSIM is the same for all methods despite DALL-E and DALL-E+VQGAN having worse PCE. This suggests CLIP \cite{clip} trained on general images is not good at spotting human body errors.

\vspace{2mm}
\subsection{Comparison with DALL-E+VQGAN} \label{sec:dalle_vqgan}
\vspace{-2mm}
Both KPE and DALL-E+VQGAN produce high-quality people images with an accurate poses. From Table \ref{table:deepfashion} against the baselines DALL-E and DALL-E+VQGAN, we can see that its OKS score matches KPE and is only marginally behind in IS and mask-SSIM, but PCE is twice the error rate of KPE. Apart from generating improved images, there are several advantages of using KPE that make it an overall superior method:
\begin{itemize}[noitemsep]
    \item \textbf{Less memory}. The keypoint token length is smaller than the pose image token, requiring less computational memory. In our experiment, the image token length is 256 while there are only 25 keypoint tokens, making it at least 10$\times$ more memory efficient.
    \item \textbf{Faster to run.} The reduction of token number reduces computational complexity, which is in $O(N^2)$ for the transformer. Furthermore, encoding pose image using VQ-VAE is computationally expensive, and removing this step can improve speed. As a result, the KPE model's inference speed is only 0.6 second on RTX5000 GPU, is 73\% faster than DALL-E+VQGAN; there is no speed penalty compared to DALL-E. 
    \item \textbf{Invariant pose representation.} The same VQ-VAE encodes both the pose and target images. However, as VQ-VAE is normally pretrained on natural images, thus the trained VQ-VAE may not perform well on synthetic skeleton images.  In contrast, KPE relies only on the keypoint information and is invariant to the image nor VQ-VAE. 
    \item \textbf{Scalable.} Increasing target image resolution or quality will require an increase in image token length hence more memory and slower running. Since KPE is invariant to the image, the pose processing will not increase computational resources as the image resolution increases. This makes our method easier to scale to higher image resolution. 
\end{itemize}

\subsection{Ablation Study and PCE} \label{sec:error}
\begin{figure}
\centering
    \begin{subfigure}{0.16\textwidth}
    \includegraphics[width=1\textwidth]{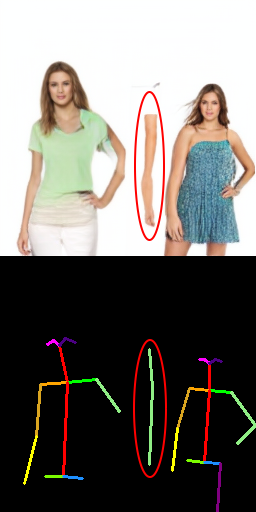}
    \caption{gt=2, h=3}
    \label{fig:error_a}
    \end{subfigure}
    \begin{subfigure}{0.16\textwidth}
    \includegraphics[width=1\textwidth]{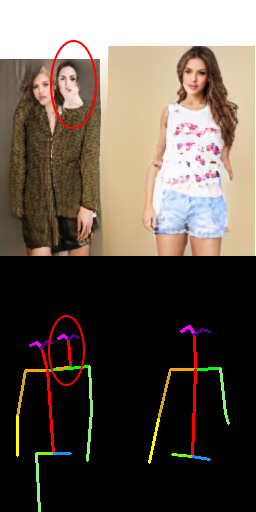}
    \caption{gt=2, h=3}
    \label{fig:error_b}
    \end{subfigure}
    \begin{subfigure}{0.16\textwidth}
    \includegraphics[width=1\textwidth]{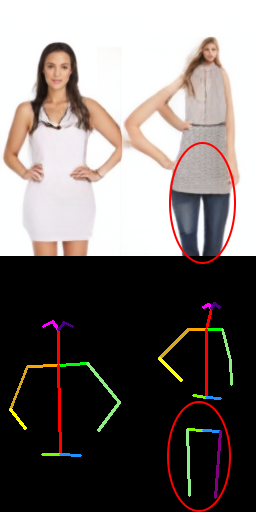}    
    \caption{gt=2, h=3}
    \label{fig:error_c}
    \end{subfigure}
    \begin{subfigure}{0.16\textwidth}
    \includegraphics[width=1\textwidth]{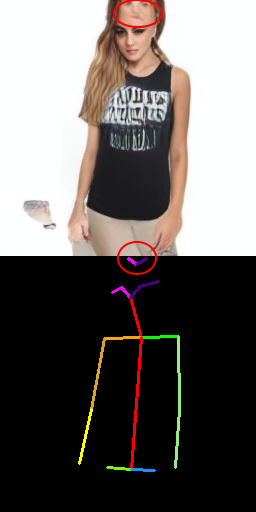}
    \caption{gt=1, h=2}
    \label{fig:error_d}
    \end{subfigure}
    \begin{subfigure}{0.16\textwidth}
    \includegraphics[width=1\textwidth]{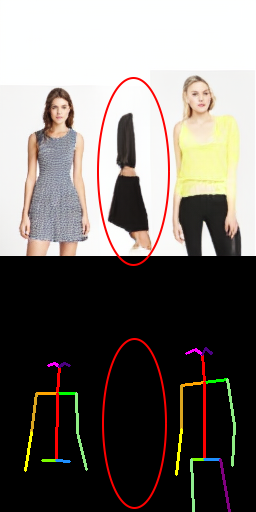}
    \caption{gt=3, h=2}
    \label{fig:error_e}
    \end{subfigure}
    \vspace{1mm}
    \caption{Top row are erroneous images generated using DALL-E, and the bottom row shows the keypoints obtained from the images. PCE can capture image errors by comparing the generated (gt) and intended (h) number of people.}
\label{fig:error}
\end{figure}

We did an ablation study comparing KPE against the baseline \cite{dalle}, a text-to-image model without pose guidance. KPE tops all the metrics in Table~\ref{table:deepfashion}, most notably with PCE rate at 0.6 $\times 10^{-3}$, which is over 13 times better than baseline's 8.2 $\times 10^{-3}$. Figure \ref{fig:error} shows an example of images that contains errors and how we can spot the error by using PCE. Figure \ref{fig:error_a} contains two realistically looking people but with an additional long arm floating in the centre of the image. The floating arm is assigned to the third person, causing PCE=1 as $h(x)\ne gt(x)$. Although measuring the discrepancy in people's count does not catch every error, it allows a standardised metric across a wide range of examples without the need for a manual visual inspection of each. Figure \ref{fig:error_b} to  \ref{fig:error_d} show examples of additional body parts where PCE=1, some of which can be difficult to see initially, like the additional face in Figure \ref{fig:error_d}. Also, we found that the baseline  \cite{dalle} sometimes generates fewer or more people than the text description. When it happens, it tends to contain some standalone body parts like Figure \ref{fig:error_a} and Figure \ref{fig:error_e}. The PCE can pick up the error in Figure \ref{fig:error_e} despite OpenPose failing to detect the incomplete person in the centre. 

\section{Limitations}
\begin{figure}[!htb]
\centering
    \begin{subfigure}{0.24\textwidth}
    \includegraphics[width=1\textwidth]{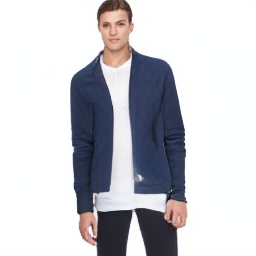}
    \caption{}
    \label{fig:limitation_a}
    \end{subfigure}
    \begin{subfigure}{0.24\textwidth}
    \includegraphics[width=1\textwidth]{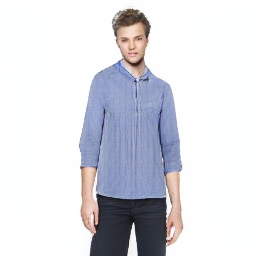}
    \caption{}
    \label{fig:limitation_b}
    \end{subfigure}
    \begin{subfigure}{0.24\textwidth}
    \includegraphics[width=1\textwidth]{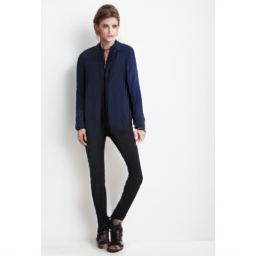}
    \caption{}
    \label{fig:limitation_c}
    \end{subfigure}
    \begin{subfigure}{0.24\textwidth}
    \includegraphics[width=1\textwidth]{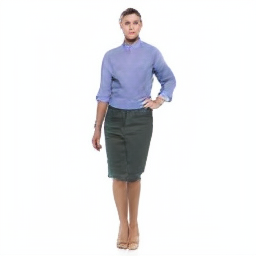}
    \caption{}
    \label{fig:limitation_d}
    \end{subfigure}
    \caption{Text of ``a man wore blue shirt'' was used to generate images from pose that is more masculine (a) towards more feminine pose in (d).}
\label{fig:limitation}
\end{figure}

The DeepFashion dataset is hugely imbalanced, where men form only a tiny fraction of the dataset, and the rest are long-haired white females. Therefore, a pose-only guided model trained on the dataset is more likely to generate female images. To understand the effect of the gender bias, we generated images of various poses using the exact text prompt  ``a man wore blue shirt'' as shown in Figure \ref{fig:limitation}.  

We can see in Figure \ref{fig:limitation_a} and \ref{fig:limitation_b} that despite the gender bias in the dataset, our model can generate convincing men from neutral poses. However, when presented with poses that are exclusive to females in the dataset, the generated images (Figure \ref{fig:limitation_c} and \ref{fig:limitation_d}) lean toward feminine appearance. This limitation implies that pose is not entirely disentangled from gender, and the model learned the gender bias from poses containing information about body proportion. Given this insight, in future, we will collect and apply our approach to a more balanced dataset in the future to address this bias in gender and ethnicity.

\section{Conclusions}
We have proposed text and pose keypoint to the generative image model creating photorealistic multiperson images. We methodologically show that adding pose as guidance improves the image quality over the SOTA text only guided generative image model. Not only does KPE generates a better image and produce a more accurate pose than existing methods, but it is also computationally efficient and does not add inference speed overhead. This allows for easy integration into transformer models. To measure the performance of generative images of people, we propose a suitable metric PCE to detect the false positive occurrence of generated people, overall creating SOTA performance both quantitatively and qualitatively.

\clearpage
%
%
\bibliography{egbib}

\begin{thebibliography}{10}

\bibitem{deepfashion}
Ziwei, Ping Luo, Shi Qiu, Xiaogang Wang, and Xiaoou~Liu Tang.
\newblock Deepfashion: Powering robust clothes recognition and retrieval with
  rich annotations.
\newblock {\em Proceedings of IEEE Conference on Computer Vision and Pattern
  Recognition (CVPR)}, 2016.

\bibitem{styletransfer}
Leon Gatys, Alexander Ecker, and Matthias Bethge.
\newblock Image style transfer using convolutional neural networks.
\newblock {\em IEEE Conference on Computer Vision and Pattern Recognition
  (CVPR)}, 2016.

\bibitem{text2human}
Yuming Jiang, Shuai Yang, Haonan Qiu, Wayne Wu, Chen~Change Loy, and Ziwei Liu.
\newblock Text2human: Text-driven controllable human image generation.
\newblock {\em SIGGRAPH}, 2022.

\bibitem{transformer}
Ashish Vaswani, Noam Shazeer, Niki Parmar, Jakob Uszkoreit, Llion Jones,
  Aidan~N. Gomez, Lukasz Kaiser, and Illia Polosukhin.
\newblock Attention is all you need.
\newblock {\em Conference on Neural Information Processing Systems (NeurIPS)},
  2017.

\bibitem{dalle}
Aditya Ramesh, Mikhail Pavlov, Gabriel Goh, Scott Gray, Chelsea Voss, Alec
  Radford, Mark Chen, and Ilya Sutskever.
\newblock Zero-shot text-to-image generation.
\newblock {\em International Conference on Machine Learning (ICML)}, 2021.

\bibitem{Makeascene}
Oran Gafni, Adam Polyak, Oron Ashual, Shelly Sheynin, Devi Parikh, and Yaniv
  Taigman.
\newblock Make-a-scene: Scene-based text-to-image generation with human priors.
\newblock {\em Arxiv preprint: 2203.13131}, 3 2022.

\bibitem{vqgan}
Patrick Esser, Robin Rombach, and Björn Ommer.
\newblock Taming transformers for high-resolution image synthesis.
\newblock {\em IEEE Conference on Computer Vision and Pattern Recognition
  (CVPR)}, 2021.

\bibitem{stackgan}
Han Zhang, Tao Xu, Hongsheng Li, Shaoting Zhang, Xiaogang Wang, Xiaolei Huang,
  and Dimitris Metaxas.
\newblock Stackgan: Text to photo-realistic image synthesis with stacked
  generative adversarial networks.
\newblock {\em International Conference on Computer Vision (ICCV)}, 2017.

\bibitem{attngan}
Tao Xu, Pengchuan Zhang, Qiuyuan Huang, Han Zhang, Zhe Gan, Xiaolei Huang, and
  Xiaodong He.
\newblock Attngan: Fine-grained text to image generation with attentional
  generative adversarial networks.
\newblock {\em IEEE Conference on Computer Vision and Pattern Recognition
  (CVPR)}, 2017.

\bibitem{dmgan}
Minfeng Zhu, Pingbo Pan, Wei Chen, and Yi~Yang.
\newblock Dm-gan: Dynamic memory generative adversarial networks for
  text-to-image synthesis.
\newblock {\em IEEE/CVF Conference on Computer Vision and Pattern Recognition
  (CVPR)}, 4 2019.

\bibitem{dfgan}
Ming Tao, Hao Tang, Fei Wu, Xiao-Yuan Jing, Bing-Kun Bao, and Changsheng Xu.
\newblock Df-gan: A simple and effective baseline for text-to-image synthesis.
\newblock 8 2020.

\bibitem{xmcgan}
Han Zhang, Jing~Yu Koh, Jason Baldridge, Honglak Lee, and Yinfei Yang.
\newblock Cross-modal contrastive learning for text-to-image generation.
\newblock {\em IEEE Conference on Computer Vision and Pattern Recognition
  (CVPR)}, 2021.

\bibitem{cgan}
Mehdi Mirza and Simon Osindero.
\newblock Conditional generative adversarial nets.
\newblock {\em Arxiv preprint:1411.1784}, 2014.

\bibitem{vqvae}
Aaron van~den Oord, Oriol Vinyals, and Koray Kavukcuoglu.
\newblock Neural discrete representation learning.
\newblock {\em Conference on Neural Information Processing Systems (NeurIPS)},
  2017.

\bibitem{cogview}
Ming Ding, Zhuoyi Yang, Wenyi Hong, Wendi Zheng, Chang Zhou, Da~Yin, Junyang
  Lin, Xu~Zou, Zhou Shao, Hongxia Yang, and Jie Tang.
\newblock Cogview: Mastering text-to-image generation via transformers.
\newblock {\em Conference on Neural Information Processing Systems (NeurIPS)},
  2021.

\bibitem{nuwa}
Chenfei Wu, Jian Liang, Lei Ji, Fan Yang, Yuejian Fang, Daxin Jiang, and Nan
  Duan.
\newblock N\"uwa: Visual synthesis pre-training for neural visual world
  creation.
\newblock {\em arxiv preprint:2111.12417}, 2021.

\bibitem{Ho2020}
Jonathan Ho, Ajay Jain, and Pieter Abbeel.
\newblock Denoising diffusion probabilistic models.
\newblock 2020.

\bibitem{Ho2021}
Tim~Ho Jonathan~Salimans.
\newblock Classifier-free diffusion guidance.
\newblock {\em NeurIPS 2021 Workshop DGMs Applications}, 2021.

\bibitem{Dhariwal2021}
Prafulla Dhariwal and Alex Nichol.
\newblock Diffusion models beat gans on image synthesis.
\newblock 2021.

\bibitem{unet}
Olaf Ronneberger, Philipp Fischer, and Thomas Brox.
\newblock U-net: Convolutional networks for biomedical image segmentation.
\newblock {\em International Conference on Medical Image Computing and Computer
  Assisted Interventions (MICCAI)}, 5 2015.

\bibitem{glide}
Alex Nichol, Prafulla Dhariwal, Aditya Ramesh, Pranav Shyam, Pamela Mishkin,
  Bob McGrew, Ilya Sutskever, and Mark Chen.
\newblock Glide: Towards photorealistic image generation and editing with
  text-guided diffusion models.
\newblock 2021.

\bibitem{dalle2}
Aditya Ramesh, Prafulla Dhariwal, Alex Nichol, Casey Chu, and Mark Chen.
\newblock Hierarchical text-conditional image generation with clip latents.
\newblock {\em Arxiv Preprint: 2204.06125}, 4 2022.

\bibitem{imagen}
Chitwan Saharia, William Chan, Saurabh Saxena, Lala Li, Jay Whang, Emily
  Denton, Seyed Kamyar~Seyed Ghasemipour, Burcu~Karagol Ayan, S.~Sara Mahdavi,
  Rapha~Gontijo Lopes, Tim Salimans, Jonathan Ho, David~J Fleet, and Mohammad
  Norouzi.
\newblock Photorealistic text-to-image diffusion models with deep language
  understanding.
\newblock {\em Arxiv preprint: 2205.11487}, 5 2022.

\bibitem{Ma2017}
Liqian Ma, Xu~Jia, Qianru Sun, Bernt Schiele, Tinne Tuytelaars, and Luc~Van
  Gool.
\newblock Pose guided person image generation.
\newblock {\em Conference on Neural Information Processing Systems (NeurIPS)},
  2017.

\bibitem{Siarohin2018}
Aliaksandr Siarohin, Enver Sangineto, Stephane Lathuiliere, and Nicu Sebe.
\newblock Deformable gans for pose-based human image generation.
\newblock {\em IEEE Conference on Computer Vision and Pattern Recognition
  (CVPR)}, 2018.

\bibitem{Yang2020}
Lingbo Yang, Pan Wang, Chang Liu, Zhanning Gao, Peiran Ren, Xinfeng Zhang,
  Shanshe Wang, Siwei Ma, Xiansheng Hua, and Wen Gao.
\newblock Towards fine-grained human pose transfer with detail replenishing
  network.
\newblock {\em IEEE Transactions on Image Processing}, 2020.

\bibitem{gaugan}
Taesung Park, Ming-Yu Liu, Ting-Chun Wang, and Jun-Yan Zhu.
\newblock Semantic image synthesis with spatially-adaptive normalization.
\newblock {\em IEEE Conference on Computer Vision and Pattern Recognition
  (CVPR)}, 2019.

\bibitem{pix2pix}
Phillip Isola, Jun-Yan Zhu, Tinghui Zhou, and Alexei~A. Efros.
\newblock Image-to-image translation with conditional adversarial networks.
\newblock {\em IEEE Conference on Computer Vision and Pattern Recognition
  (CVPR)}, 2016.

\bibitem{Zhu2019}
Zhen Zhu, Tengteng Huang, Baoguang Shi, Miao Yu, Bofei Wang, and Xiang Bai.
\newblock Progressive pose attention transfer for person image generation.
\newblock {\em IEEE Conference on Computer Vision and Pattern Recognition
  (CVPR)}, 2019.

\bibitem{Tompson2014}
Jonathan Tompson, Arjun Jain, Yann LeCun, and Christoph Bregler.
\newblock Joint training of a convolutional network and a graphical model for
  human pose estimation.
\newblock {\em NeurIPS,}, 2014.

\bibitem{dalle_mini}
Boris Dayma, Suraj Patil, Pedro Cuenca, Khalid Saifullah, Tanishq Abraham, Phuc
  Le~Khac, Luke Melas, and Ritobrata Ghosh.
\newblock Dall·e mini, 2021.

\bibitem{skeletor}
Tao Jiang, Necati~Cihan Camgoz, and Richard Bowden.
\newblock Skeletor: Skeletal transformers for robust body-pose estimation.
\newblock {\em IEEE Conference on Computer Vision and Pattern Recognition
  Workshops (CVPRW)}, 2021.

\bibitem{Li2021}
Ke~Li, Shijie Wang, Xiang Zhang, Yifan Xu, Weijian Xu, and Zhuowen Tu.
\newblock Pose recognition with cascade transformers.
\newblock {\em IEEE Conference on Computer Vision and Pattern Recognition
  (CVPR)}, 2021.

\bibitem{axial}
Jonathan Ho, Nal Kalchbrenner, Dirk Weissenborn, and Tim Salimans.
\newblock Axial attention in multidimensional transformers.
\newblock {\em arxiv preprint:1912.12180v1}, 2019.

\bibitem{dalle_pytorch}
Phil Wang.
\newblock https://github.com/lucidrains/dalle-pytorch, 2021.

\bibitem{bpe}
Rico Sennrich, Barry Haddow, and Alexandra Birch.
\newblock Neural machine translation of rare words with subword units.
\newblock {\em Association for Computational Linguistics (ACL)}, 2015.

\bibitem{vit}
Alexey Dosovitskiy, Lucas Beyer, Alexander Kolesnikov, Dirk Weissenborn,
  Xiaohua Zhai, Thomas Unterthiner, Mostafa Dehghani, Matthias Minderer, Georg
  Heigold, Sylvain Gelly, Jakob Uszkoreit, and Neil Houlsby.
\newblock An image is worth 16x16 words: Transformers for image recognition at
  scale.
\newblock {\em International Conference for Learning Representations (ICLR)},
  2020.

\bibitem{openpose}
Zhe Cao, Gines Hidalgo, Tomas Simon, Shih-En Wei, and Yaser Sheikh.
\newblock Openpose: Realtime multi-person 2d pose estimation using part
  affinity fields.
\newblock {\em IEEE Transactions on Pattern Analysis and Machine Intelligence},
  2019.

\bibitem{oks}
MSCOCO.
\newblock https://cocodataset.org/, 2017.

\bibitem{godiva}
Chenfei Wu, Lun Huang, Qianxi Zhang, Binyang Li, Lei Ji, Fan Yang, Guillermo
  Sapiro, and Nan Duan.
\newblock Godiva: Generating open-domain videos from natural descriptions.
\newblock {\em Arxiv Preprint 2104.14806}, 4 2021.

\bibitem{is}
Tim Salimans, Ian Goodfellow, Wojciech Zaremba, Vicki Cheung, Alec Radford, and
  Xi~Chen.
\newblock Improved techniques for training gans.
\newblock {\em arXiv:1606.03498}, 2016.

\bibitem{fid}
Martin Heusel, Hubert Ramsauer, Thomas Unterthiner, Bernhard Nessler, and Sepp
  Hochreiter.
\newblock Gans trained by a two time-scale update rule converge to a local nash
  equilibrium.
\newblock {\em Advances in Neural Information Processing Systems 30 (NIPS
  2017)}, 2017.

\bibitem{is_note}
Shane Barratt and Rishi Sharma.
\newblock A note on the inception score.
\newblock {\em International Conference on Machine Learning (ICML)}, 2018.

\bibitem{body25}
OpenPose.
\newblock
  https://cmu-perceptual-computing-lab.github.io/openpose/web/html/doc/md\_doc\_02\_output.html,
  2020.

\bibitem{adam}
Diederik~P. Kingma and Jimmy Ba.
\newblock Adam: A method for stochastic optimization.
\newblock {\em International Conference for Learning Representations (ICLR)},
  2014.

\bibitem{clip}
Alec Radford, Jong~Wook Kim, Chris Hallacy, Aditya Ramesh, Gabriel Goh,
  Sandhini Agarwal, Girish Sastry, Amanda Askell, Pamela Mishkin, Jack Clark,
  Gretchen Krueger, and Ilya Sutskever.
\newblock Learning transferable visual models from natural language
  supervision.
\newblock {\em International Conference on Machine Learning (ICML)}, 2 2021.

\end{thebibliography}
\end{document}